\documentclass[journal,web]{article}
\usepackage{tmi}
\usepackage{cite}
\usepackage{amsmath,amssymb,amsfonts}
\usepackage{algorithmic}
\usepackage{graphicx}
\usepackage{textcomp}
\usepackage{xspace}
\usepackage{amssymb}
\usepackage{tabularx, booktabs}
\usepackage{booktabs}
\usepackage{amsmath}
\usepackage{array} 
\usepackage{picture}
\usepackage[pagebackref=true,breaklinks=true,colorlinks,bookmarks=false]{hyperref}
\usepackage{backrefbis}
\usepackage{color}
\usepackage[algo2e,ruled,vlined]{algorithm2e}
\usepackage{amsmath}
\usepackage{amssymb}
\usepackage{hyperref}
\usepackage[]{algorithm}
\usepackage{comment}
\usepackage{multirow}

\usepackage[margin=1in]{geometry}
\usepackage{changepage}
\newcommand*{\colorboxed}{}
\def\colorboxed#1#{%
 \colorboxedAux{#1}%
}
\newcommand*{\colorboxedAux}[3]{%
 \begingroup
 \colorlet{cb@saved}{.}%
 \color#1{#2}%
 \boxed{%
 \color{cb@saved}%
 #3%
 }%
 \endgroup
}

\usepackage{tikz}
\usepackage{textcomp}
\usepackage{hyperref}
\usepackage{lipsum}

\newcommand\copyrighttext{%
  \footnotesize \textcopyright 1558-254X © 2021 IEEE. Personal use is permitted, but republication/redistribution requires IEEE permission.
See https://www.ieee.org/publications/rights/index.html for more information. 
  DOI: \href{<http://tex.stackexchange.com>}{10.1109/TMI.2021.3067688}}
\newcommand\copyrightnotice{%
\begin{tikzpicture}[remember picture,overlay]
\node[anchor=south,yshift=10pt] at (current page.south) {\fbox{\parbox{\dimexpr\textwidth-\fboxsep-\fboxrule\relax}{\copyrighttext}}};
\end{tikzpicture}%
}

\usepackage{array}
\newcolumntype{L}[1]{>{\raggedright\let\newline\\\arraybackslash\hspace{0pt}}m{#1}}
\newcolumntype{C}[1]{>{\centering\let\newline\\\arraybackslash\hspace{0pt}}m{#1}}
\newcolumntype{R}[1]{>{\raggedleft\let\newline\\\arraybackslash\hspace{0pt}}m{#1}}

\newcommand{\beq}{\begin{equation}}
\newcommand{\eeq}{\end{equation}}

\def\BibTeX{{\rm B\kern-.05em{\sc i\kern-.025em b}\kern-.08em
    T\kern-.1667em\lower.7ex\hbox{E}\kern-.125emX}}
\begin{document}
\title{Constrained Domain Adaptation for~Image~Segmentation}
\author{M. Bateson, J. Dolz, H. Kervadec, H. Lombaert, I. Ben Ayed

\thanks{This work is supported by the Natural Sciences and Engineering Research Council of Canada (NSERC), Discovery Grant program, by The Fonds de recherche du Québec - Nature et technologies (FRQNT) grant, the Canada Research Chair on Shape Analysis in Medical Imaging, the ETS Research Chair on Artificial Intelligence in Medical Imaging, and NVIDIA with a donation of a GPU. 
}
\thanks{M. Bateson (e-mail: mathildebateson@gmail.com), J.Dolz, H. Kervadec, H. Lombaert, I. Ben Ayed are with Ecole de Technologie Supérieure, Montréal, Canada}}

\maketitle
\copyrightnotice{}
\begin{abstract}
Domain Adaption tasks have recently attracted substantial attention in computer vision as they improve the transferability of deep network models from a source to a target domain with different characteristics. A large body of state-of-the-art domain-adaptation
methods was developed for image classification purposes, which may be inadequate for segmentation tasks. We propose to adapt segmentation networks with a constrained formulation, which embeds domain-invariant prior knowledge about the segmentation regions. Such knowledge may take the form of anatomical information, for instance, structure size or shape, which can be known a priori or learned from the source samples via an auxiliary task. Our general formulation imposes inequality constraints on the network predictions of unlabeled or weakly labeled target samples, thereby matching implicitly the prediction statistics of the target and source domains, with permitted uncertainty of prior knowledge. Furthermore, our inequality constraints easily integrate weak annotations of the target data, such as image-level tags. We address the ensuing constrained optimization problem with differentiable penalties, fully suited for conventional stochastic gradient descent approaches. Unlike common two-step adversarial training, our formulation is based on a single segmentation network, which simplifies adaptation, while improving training quality. Comparison with state-of-the-art adaptation methods reveals considerably better performance of our model on two challenging tasks. Particularly, it consistently yields a performance gain of 1-4\% Dice across architectures and datasets. Our results also show robustness to imprecision in the prior knowledge. The versatility of our novel approach can be readily used in various segmentation problems, with code available publicly.
\end{abstract}


\section{Introduction}
\label{sec:introduction}
Building accurate automatic image analysis systems is a key problem for many biomedical applications. In recent years, Convolutional Neural Networks (CNN) have made a substantial impact and became the de-facto choice in a breadth of computer vision and medical imaging tasks, including classification, detection, semantic segmentation, \cite{litjens2017survey,DolzNeuro2017}, achieving state-of-the-art or even human-level performances. Nonetheless, CNN typically require a huge quantity of annotated data to perform well. This is especially the case for semantic segmentation, an important first step of the diagnosis and treatment pipeline. To train CNNs for segmentation, pixel or voxel-level annotations of large datasets are commonly used. In 3D medical images, such voxel-level annotations are very cumbersome to obtain, as they require scarce expert knowledge. This has led to many recent efforts, both in computer vision \cite{Dai2015,Pinheiro2014FromIT,Wei2016STCAS} and medical imaging \cite{Rajchl2016,kervadec2019constrained,Jia2017}, to develop methodologies mitigating the lack of full annotations, such as semi- or weakly-supervised models \cite{Bai,Zhou, Dai2015,Tang2018NormalizedCL}. 

\begin{figure}[t]
 \begin{center}
 \mbox{
 \includegraphics[width=0.7\linewidth]{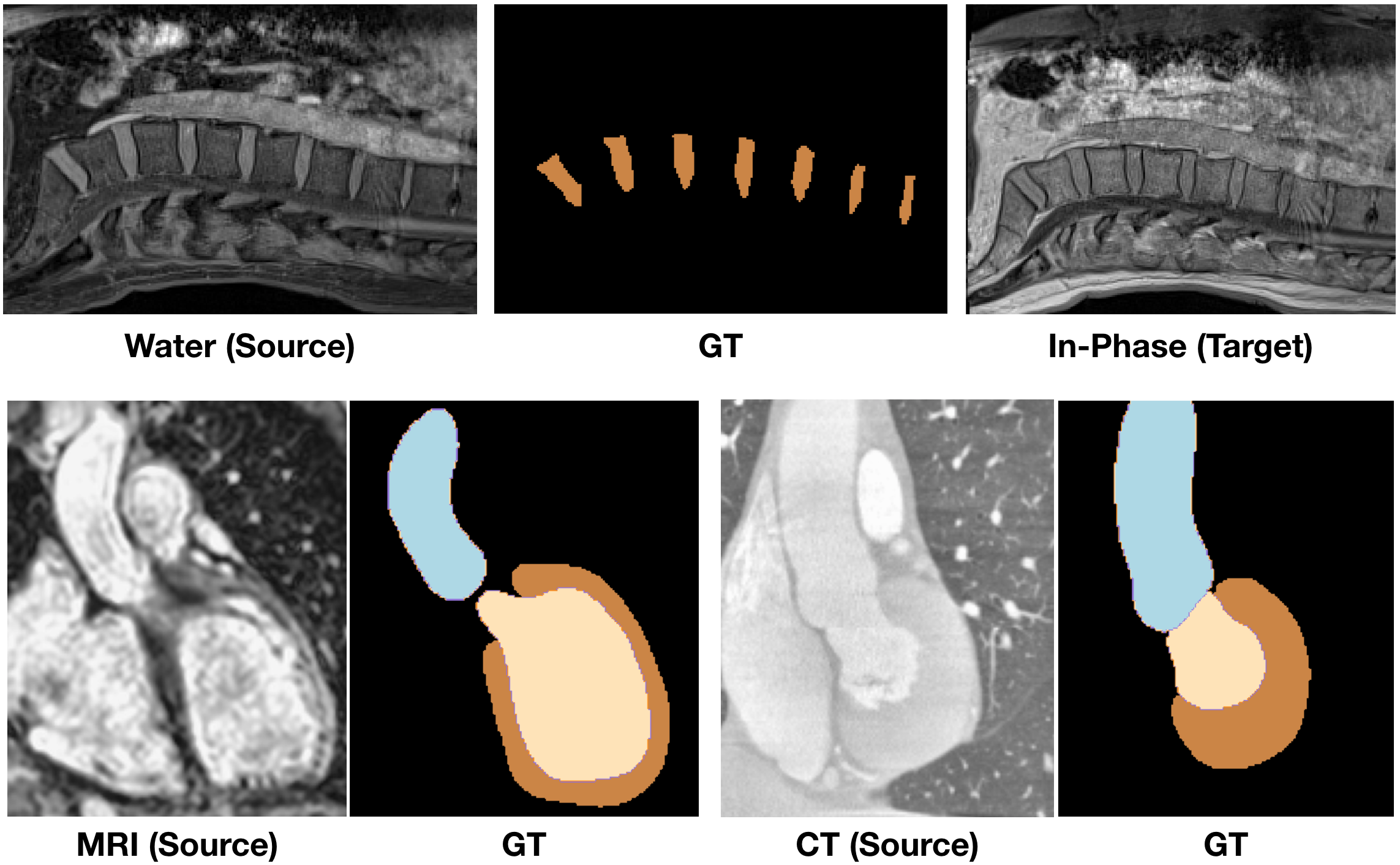}
 }
\caption{Visualization of severe domain shifts between source and target modalities in two applications. Top: 2 aligned spine images from Water and In-Phase MRI and the corresponding ground-truth segmentation, with the intervertebral disks depicted in brown and the background in black. Bottom: 2 cardiac images from MRI and CT, and their ground-truth segmentations. The cardiac structures of AA, LVC and and MYO are depicted in blue, purple and brown, respectively.}
\label{fig:s_t_im}
\end{center}
\end{figure}

Typically, segmentation ground-truth is available for very limited data, and the performances of supervised models might drop significantly with new unlabeled samples (target data) that differ from the labeled training samples (source data). These under-performances are a major drawback of standard deep learning techniques, which impedes their deployment in practical scenarios with domain shifts. 
In medical imaging, such domain-shift scenarios occur frequently when acquisition machines come from different vendors and clinical sites, or when images are acquired across multiple protocols, such as Computed Tomography (CT) and Magnetic Resonance Imaging (MRI) modalities. Acquiring images across different modalities is often very useful to capture different physical properties, which play complementary roles in the clinical procedures for disease diagnosis and treatment. In Figure \ref{fig:s_t_im}, the lower spine is shown across two distinct MRI modalities (Water and In-Phase), and the heart across MRI and CT. In both cases, one can observe significant differences in contrast, intensity histograms and demarcation between the structures in the two modalities.

Domain adaptation (DA) techniques aim at learning models robust to such distribution shifts. Early deep-learning approaches for domain adaptation investigated minimizing a distance between the features learned from the source and target domains, e.g., the Maximum Mean Discrepancy (MMD) work in \cite{Tzeng2014DeepDC}, thereby 
aligning the source and target distributions in the latent feature space. 
With the recent success of generative adversarial networks (GANs) \cite{Goodfellow}, adversarial learning has become the dominating choice for domain adaptation \cite{hoffman2017cycada,Bousmalis2017,Russo2018, sankaranarayanan2017generate,Chen2018}.

Unlike adversarial methods, we introduce our Constrained Domain Adaptation to guide the network learning with domain knowledge, e.g., anatomical or 
learned priors. By enforcing inequality constraints on the network output in the target domain, our method enables to implicitly match prediction statistics between source and target domains, without the burden of two-step adversarial training such as in GANs. Moreover, based on inequality constraints, our 
framework allows uncertainty in the domain knowledge and can therefore leverage weak labels of the target samples, for instance, in the form of image-level tags for segmentation tasks.

\subsection{Related work}
\label{ssec:prior}

Domain adaptation is currently attracting substantial research efforts, both in computer vision \cite{hoffman2017cycada,tsai2018learning,ADDA} and medical imaging \cite{Cheplygina2018Notsosupervised,kamnitsas2017unsupervised,ren2018adversarial,zhang2018task}. The first attempts to tackle domain shifts were proposed in the vision community, for classification tasks\cite{ADDA,shu2018dirt}. When the labels are not available for the target domain, the problem is referred to as unsupervised domain adaptation (UDA), and is often formulated as a domain-divergence minimization. These methods rely on the minimization of a discrepancy between distributions, and can be performed at various levels. For instance, they could be deployed in the input space, transforming the images from the source domain so that they look `'similar'' to those from the target domain, or vice-versa. This approach has enabled a new line of works in both computer vision \cite{hoffman2017cycada,Bousmalis2017,Russo2018, sankaranarayanan2017generate} and medical imaging \cite{Chen2018}, but still remains inadequate when the domain shift is too large, which may limit its broad applicability. Such a discrepancy minimization  could also operate on the representation learned by the CNN. This amounts to aligning the intermediate features from both domains \cite{Ghifary,kamnitsas2017unsupervised,long2015learning,Long2016}, potentially helping the learned representation to be both useful for classification and invariant with respect to domain shifts \cite{ChenMIDL}. The most common discrepancy minimization technique uses an adversarial formulation: a discriminator tries to distinguish between the source and the target domain, while the classifier performs the original task of classifying images from both domains. 

Beyond image classification tasks, for which excellent performances were reported \cite{ADDA,shu2018dirt}, there is a rapidly growing interest in adapting segmentation networks \cite{Opbroek,kamnitsas2017unsupervised,tsai2018learning}, as building pixel-level labels for each new domain is even more tedious. 
A recent body of work extended domain adaptation ideas for the segmentation of images from different domains \cite{TAJBAKHSH2020}. Most of the studies adapting segmentation networks, either for medical \cite{gholami2018novel,javanmardi2018domain,kamnitsas2017unsupervised,zhao2019supervised,Khalili} or natural images \cite{chen2018road,hoffman2017cycada,hong2018conditional,tsai2018learning} use adversarial training. The latter alternates the training of two networks, one learning a discriminator between source and target features and the other generating segmentations. 
However, the dimensionality of the learned features and the label space in a segmentation task is much higher than in classification tasks. This might invalidate the assumption that the source and target share the same representation at all the abstraction levels of a deep network. To address this problem, Tsai \textit{et al.} \cite{tsai2018learning} proposed to minimize the discrepancy in the softmax-output space, outperforming feature-matching techniques for unsupervised adaptation in the context of color image segmentation. The underlying motivation is that the output space conveys domain-invariant information about segmentation structures, for instance, shape and spatial layout, even when the inputs across domains are substantially different. 

Despite the clear benefits of adversarial techniques in DA for classification \cite{ADDA}, our experiments suggest that adversarial training may not be well suited to adapt segmentation networks. As pointed out in a few recent work in computer vision \cite{zhang2017curriculum,zou2018unsupervised}, learning a discriminator boundary between the source and target domains is much more complex for segmentation, as it involves predictions in an exponentially large label space. Instead, it has been shown that self-training \cite{zhang2017curriculum}, which generates masks of unlabeled target images via the network’s own predictions and uses priors on the spatial layout of the segmentation regions, can yield better performances. In an approach related to our work \cite{zhang2017curriculum}, but investigated for color images, the authors showed that a curriculum learning strategy, which minimizes a Kullback–Leibler (KL) divergence between image-level distributions, for instance, region proportions, can be more effective than adversarial techniques. Finally, it is worth mentioning the recent classification study in \cite{shu2018dirt}, which argued that adversarial training is not sufficient for high-capacity models, as is the case for segmentation. For deep architectures, the authors of \cite{shu2018dirt} showed experimentally that jointly minimizing source generalization error and feature divergence does not yield high accuracy on the target task.

Our study draws upon several recent weakly- and semi-supervised segmentation work \cite{kervadec2019constrained,Jia2017,KervadecMiccai,Tang2018NormalizedCL,Tang2018Regularized,Pathak2015Constrained}, which imposed regularization or prior-knowledge terms on the predictions of deep networks, leveraging unlabeled or weakly labeled data. For instance, the work in \cite{Tang2018NormalizedCL,Tang2018Regularized} showed that regularization losses, in the form of a dense conditional random field (CRF) or a balanced graph clustering loss, can achieve excellent segmentation results using only a small fraction of labeled pixels, approaching full-supervision performances in the context of colour images. Along this same vein of research, the work in \cite{kervadec2019constrained,Jia2017} incorporated priors on the sizes of the segmentation regions via additional loss functions. In this weakly or semi-supervised setting, the main assumption is that the unlabeled data is assumed to be drawn from the same distribution as the labeled data (i.e., there are no domain shifts), which makes the task less challenging than unsupervised or weakly supervised domain adaptation.

\subsection{Contributions}
We propose a general Constrained Domain Adaptation (CDA) formulation for semantic segmentation, which embeds domain-invariant prior knowledge about the segmentation regions. In medical imaging, such knowledge may take the form of simple anatomical information, for instance, structure size or shape, which can be either known {\em a priori} or learned from the source samples via an auxiliary task. Our general formulation imposes inequality constraints on the network predictions of unlabeled or weakly labeled target samples, thereby matching implicitly the prediction statistics of the target and source domains, with permitted uncertainty of prior knowledge. Furthermore, our inequality constraints enable to leverage weak annotations of the target data, for instance, in the form of simple image-level tags. We address the ensuing constrained optimization problem with differentiable penalties, fully suited for conventional stochastic gradient descent approaches. Unlike current two-step adversarial training methods, our formulation is based on a single segmentation network, which simplifies adaptation by avoiding extra adversarial steps, while improving 
training quality.

We report comprehensive evaluations and comparisons on two public segmentation challenges: the intervertebral-disc MICCAI IVD 2018 and the cardiac substructure segmentation MMWHS 2017 challenge. For the adaptation of a segmentation network from one modality to another, our proposed inequality-constrained formulation yielded significant improvements over the state-of-the-art adversarial domain adaptation method in \cite{tsai2018learning}. Moreover, CDA is much faster than adversarial techniques, as the constraints can be learned offline, while the adaptation phase only requires the training of a single segmentation network. The benchmark provided by \cite{dou2018pnp} shows that our formulation outperforms all state-of-the art adversarial methods on the cardiac dataset, including the one proposed in \cite{dou2018pnp}. We further provide a comprehensive experimental analysis of CDA, which confirms its robustness to imprecision in the prior-knowledge information. First, we showed that prior knowledge at the image level, for instance, region size, can be learned and estimated via an auxiliary network.Second, we showed that region size could also be estimated from statistics from the source domain, approaching textbook anatomical knowledge. While these estimations are uncertain, we obtained very competitive results of our segmentation network constrained with such priors.  Indeed, our method outperforms the recent curriculum adaptation method in \cite{zhang2017curriculum}, which does not allow uncertainty in the prior knowledge, in both applications. In addition to our contributions, our code is publicly available\footnote{\url{https://github.com/mathilde-b/CDA}}.

A preliminary conference version of this work appeared at MICCAI 2019 \cite{Bateson2019}. This journal version provides (1) a broader treatment of the subject with a more detailed description of the method; (2) a new application, the adaptation of cardiac substructure segmentation between MRI and CT images (3) new ablation studies that demonstrate the practical usefulness and robustness of CDA with respect to uncertainty in the prior knowledge, as well as its application to a weak-supervision setting. In particular, we performed comprehensive evaluations for the more realistic and broader setting where prior constraints about the target region are not known/precise, but rather estimated via (a) an auxiliary network, and (b) derived from source statistics, with substantial imprecision. 

\section{Methodology}

\begin{figure*}[t!]
\mbox{ \small 
 \includegraphics[width=1\linewidth]{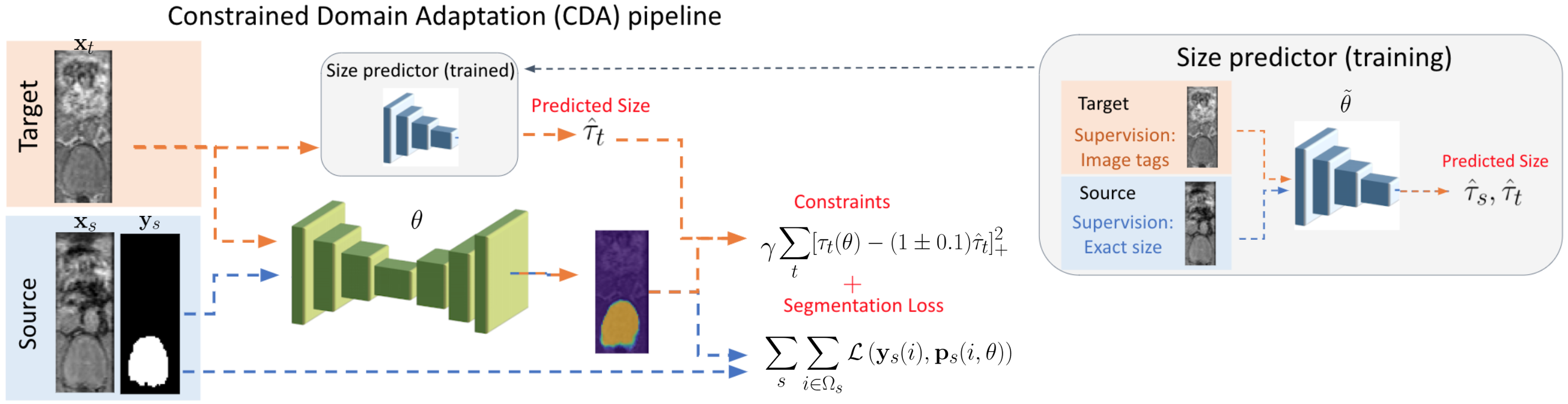} 
 }
 \vspace{-7mm}
 \caption{(\textit{Left}) Pipeline of the proposed CDA framework. The prior knowledge can be learned and predicted with an auxiliary regression network. (\textit{Right}) The training process of the auxiliary regression network.}
\label{fig:pipeline}
\end{figure*}

\subsection{The proposed Constrained Domain Adaptation}
\label{sec:CDA}
Let us denote ${I}_s: \Omega_s\subset \mathbb{R}^{2,3} \rightarrow \mathbb{R}$, $s=1, \dots, S$, as the training images of the source domain, each of them having a corresponding ground-truth segmentation, which, for each pixel (or voxel) $i \in \Omega_s$, takes the form of binary simplex vector ${\mathbf y}_s (i) = (y^1_s (i), \dots, y^K_s (i)) \in \{0,1\}^K$, with $K$ the number of classes, or segmentation regions. If we now consider $T$ images from the target domain, ${I}_t: \Omega_t\subset \mathbb R^{2,3} \rightarrow {\mathbb R}$, $t=1, \dots, T$, we can state domain adaptation for segmentation as the following 
constrained optimization problem with respect to the network parameters $\theta$:  
\begin{equation}\label{eq:constrained-domain-adaption}
\begin{aligned}
  &\min_{\theta}\sum_{s} \sum_{i \in \Omega_s} {\cal L}({\mathbf y}_s (i), {\mathbf p}_s (i, \theta))
\\
  &\text{s.t.} \quad f_c({\mathbf P}_t(\theta)) \leq 0 \quad  c = 1, \dots, C; t = 1, \dots, T
  \end{aligned}
\end{equation}
where ${\mathbf p}_x (i, \theta) = (p^1_x (i,\theta), \dots, p^K_x (i, \theta)) \in [0,1]^K$ is the softmax output of the network at pixel/voxel $i$ in image $x \in \{t=1, \dots, T\} \cup \{s=1, \dots, S \}$, and ${\mathbf P}_x(\theta)$ is a $K \times |\Omega_x|$ matrix whose columns are the vectors of network outputs ${\mathbf p}_x (i, \theta), i \in \Omega_x $. 
In problem \eqref{eq:constrained-domain-adaption}, $\cal L$ is a standard supervised-learning loss defined solely over the source data (both images and ground-truth masks), e.g., 
the cross-entropy:
\begin{equation}
{\cal L}({\mathbf y}_s (i), {\mathbf p}_s (i, \theta)) = - \sum_k y^k_s (i) \log p^k_s (i, \theta)
\end{equation}
$f_c$ denote the functions that embed constraints on unlabeled or weakly-labeled target-domain images. 

Embedding prior knowledge via inequality constraints imposed on the network outputs for target-domain data can be very practical. Assume, for instance, that we have prior knowledge about the size (or cardinality) of the target segmentation region (or class) $k$. Inequality constraints allow imprecision (or uncertainty) in this knowledge, in the form of lower and upper bounds on region size, unlike \cite{zhang2017curriculum,Jia2017,Boykov_2015_ICCV}.
For instance, when we have an upper bound $a$ on the size of region $k$, we can impose the following inequality constraint on the network outputs: 
\begin{equation}
    \label{eq_a} \sum_{i \in \Omega_t} p^k_t (i, \theta) - a \leq 0
\end{equation}
In this case, the corresponding constraint $c$ in the general-form constrained problem in Eq. \eqref{eq:constrained-domain-adaption} uses the following particular function:
\begin{equation}
\label{constraint-upper-bound}
f_c({\mathbf P}_t(\theta)) = \sum_{i \in \Omega_t}  p^k_t (i, \theta) - a
\end{equation}
Similarly, we can impose a lower bound $b$ on the size of region $k$ by using the following function instead:
\begin{equation}
\label{constraint-lower-bound}
f_c({\mathbf P}_t(\theta)) = b - \sum_{i \in \Omega_t} p^k_t (i, \theta)
\end{equation}
Our framework can be easily extended to more descriptive constraints, e.g., invariant shape moments \cite{Klodt}, which do not change from one modality to another\footnote{In fact, region size is the $0$-order shape moment; one can use higher-order shape moments for richer descriptions of shape.}.

An advantage of our formulation is that it can easily integrate weak supervision taking the form of image-level annotations, or tags, in the target domain.
Such weak annotations indicate whether a segmentation region $k$ is present or absent in a given image and, therefore, are much less time consuming than full supervision of segmentation, which requires a label for each pixel. Observe that image-level weak supervision can be written 
conveniently with an inequality constraint in the case of a negative image that does not contain target region $k$: 
\[\sum_{i \in \Omega_t} p^k_t (i, \theta) \leq 0\]
Similarly, for a positive image containing the target region $k$, we can impose the following constraint:
\[\sum_{i \in \Omega_t} p^k_t (i, \theta) > 0 \]

Imposing constraints is common in non-deep image analysis, as it allows to incorporate many types of prior knowledge, such as geometry, context or texture, 
and has proven effective in many applications, including medical image segmentation \cite{Cremers2006,Klodt,Nosrati2016IncorporatingPK}. However, with non-convex deep segmentation models, even when the constraints are convex with respect to the network probability outputs, the general problem in \eqref{eq:constrained-domain-adaption} is challenging. 
In standard convex-optimization problems, a common technique to deal with hard inequality constraints relies on the minimization of the corresponding Lagrangian dual, solving primal 
and dual problems in an alternating scheme \cite{Bertsekas1995}. For the problem in \eqref{eq:constrained-domain-adaption}, this would involve alternating the optimization of a CNN for the primal with stochastic optimization, e.g., SGD, and projected gradient-ascent iterates for the dual. For semantic segmentation networks involving millions of parameters, this might be computationally intractable. Moreover, the interplay between the primal and dual optimization in the context of deep CNNs might lead to instabilities, seriously affecting the performances of Lagrangian-dual optimization. Therefore, and as pointed out in several recent works \cite{Pathak2015Constrained,Marquez-Neila2017,kervadec2019constrained}, despite the clear benefits of imposing hard constraints on CNNs in various applications and problems, such a standard Lagrangian-dual optimization is avoided in the context of modern deep networks.

Instead, in deep networks, equality or inequality constraints are typically handled in a `'soft'' manner by augmenting the loss with a {\em penalty} function \cite{He2017,kervadec2019constrained,Jia2017}. The penalty-based approach is a simple alternative to Lagrangian optimization, and is well-known in the general context of constrained optimization; see \cite{Bertsekas1995}, Chapter 4. In general, these penalty-based methods approximate a constrained minimization problem with an unconstrained one by adding a term, which increases when the constraints are violated. A disadvantage of the penalty-based approach is that, contrary to Lagrangian optimization, it does not guarantee that the constraints will be satisfied.
But it is convenient for deep networks because it removes the requirement for explicit Lagrangian-dual optimization. The inequality constraints are fully integrated with stochastic optimization, as in standard unconstrained losses. This optimization avoids gradient ascent projections over the dual variables, and reduces the computational cost for training. Therefore, in this work, we use a penalty approach, and replace the constrained problem in \eqref{eq:constrained-domain-adaption} by the following unconstrained one: 
\begin{equation}
\label{eq:unconstrained-problem-penalty}
\min_{\theta} \sum_{s} \sum_{i \in \Omega_s} {\cal L}({\mathbf y}_s (i), {\mathbf p} (i, \theta)) + \gamma {\cal F}(\theta)
\end{equation}
where $\gamma$ is a positive constant and ${\cal F}$ a penalty, which takes the following form for the inequality constraints in \eqref{eq:constrained-domain-adaption}:
\begin{equation}
\label{eq:quadratic_penalty}
{\cal F}(\theta) = \sum_{c=1}^{C} \sum_{t=1}^{T} [f_c({\mathbf P}_t(\theta))]_+^2
\end{equation}
with $[x]_+ = \max (0,x)$ denoting the rectifier linear unit function. Clearly, when a constraint is violated, the penalty function is strictly positive; the further we get from the constraint satisfaction boundary, the larger the penalty. A satisfied constraint corresponds to a null cost.  

\subsection{Learning the constraints}
\label{sec:constraints}

An important question is how to best derive useful constraints for CDA. Depending on the application, such priors may be obtained from domain or contextual knowledge, such as anatomical knowledge for medical imaging. For instance, in the first application we tackle in our experiments, we can use human spine measurements that are well known in the clinical literature \cite{Berry87} for constraining the sizes of the intervertebral discs in axial MRI slices. When priors are invariant across domains, as is the case for the size of a segmentation region, the statistics of the priors in the source domain can be used. Another option is to train an auxiliary network to learn an 
estimation of the prior constraints. In our applications below, an auxiliary regression network is trained on the images $I_s$ from the source domain $S$, where the ground 
truth size $\tau_s$ is known. Then, the learned model is used to predict region-size constraints on the target images. Of course, this might lead to errors in size-constraint predictions over the target images due to the domain shift between the source images used for learning and the target images for inference. 
To help the regression network, one can add the images $I_t$ from the target domain, using the following "fake" sizes labels for each structure k:
 \begin{equation}
\tau_{t}=\begin{cases}
    \bar{\tau}_S & \text{if region k is within image $t$}.\\
    0 & \text{otherwise}.
  \end{cases}
  \label{eq:fakesizes}
\end{equation}
where $\bar{\tau}_S$ is the median of ground truth sizes for structure $k$ in the source images $I_s$. This corresponds to a weakly-supervised setting, where the image-level tag information in the target domain is available. 
In our experiments, the auxiliary regression network $R$ with parameters $\tilde{\boldsymbol{\theta}}$ is trained to predict the segmentation-region size $\tau_x$, within an image $x$, with the following squared $\mathcal{L}_{2}$ loss:

\begin{equation}
    \min _{\tilde{\boldsymbol{\theta}}} \sum_{x \in \mathcal{S}\cup T }\left(R(x | \tilde{\boldsymbol{\theta}})-\tau_x\right)^{2}
    \label{eq:l2}
\end{equation}

As we will see in our experiments, our inequality-constraint formulation is robust to imprecision in the prior-knowledge information. While the size prior learned and predicted via an auxiliary network is noisy (uncertain), we obtained very competitive results of our segmentation network constrained with such a prior. We recapitulate the proposed constrained domain adaptation pipeline in Figure \ref{fig:pipeline}.

\section{Experiments and results}\label{sec:resexperiments}

\subsection{Experiments set-up}\label{sec:experiments}

\subsubsection{\textbf{Dataset}}
\label{sssec:dataset}
\paragraph{\textbf{IVDM3Seg}}The proposed CDA method is first evaluated on the dataset from the MICCAI 2018 IVDM3Seg Challenge\footnote{https://ivdm3seg.weebly.com/}, a study investigating intervertebral discs (IVD) degeneration. This dataset contains 16 3D multi-modal magnetic resonance (MR) scans of the lower spine, with their corresponding manual segmentations. The 8 subjects were scanned with a 1.5-Tesla MRI Scanner from Siemens using Dixon protocol, at two different stages. In our experiments, models are trained on fully annotated volumes from the Water modality (source domain $S$), and validated on the In-Phase modality (target domain $T$). In this setting, the different MRI modalities are acquired from the same patient.
To reproduce a more realistic scenario, we have considered that the source and target images are not aligned. This contrasts with the experiments in \cite{Bateson2019}, where source and target images were registered. From this dataset, 13 scans were used for training, and the remaining 3 scans for validation. As the constraints are imposed axial-wise, we resampled the 36 coronal slices of size 256$\times$256 pixels into 256 slices of 256$\times$36 pixels, as shown in Figure \ref{fig:pipeline}. Images were normalized between 0 and 1. No additional pre-processing or data augmentation was performed. 

\paragraph{MMWHS}
We employed the 2017 Multi-Modality Whole Heart Segmentation (MMWHS) Challenge dataset for cardiac segmentation \cite{Zhuang2019}. The data consist of  20 MRI (source domain $S$) and 20 CT volumes (target domain $T$) of non-overlapping subjects, with ground-truth masks being provided for both modalities. We aim to adapt the segmentation network for parsing four cardiac structures: the ascending aorta (AA), the left atrium blood cavity (LA), the left ventricle blood cavity (LV) and the myocardium of the left
ventricle (MYO). We employed the pre-processed data provided by \cite{dou2018pnp}, as well as their data split, with 16 subjects used for training and validation and 4 for testing. In order to obtain a similar field of view for all volumes, they cropped the original scans to center the structures to segment. For each modality, a 3D bounding box with a fixed coronal plane size of $256\times256$ centered at the heart was used to crop each volume. In \cite{dou2018pnp}, a data augmentation based on affine transformations was employed on both the source and target domains for the benchmark (\textit{NoAdap}, \textit{DANN}, \textit{ADDA}, \textit{CycleGAN}, \textit{Pnp-AdaNet}). Similarly, for all the methods we implemented (\textit{NoAdap}, \textit{Adversarial}\cite{tsai2018learning}, \textit{KLAdap}\cite{zhang2017curriculum}, \textit{ConstraintAdap}, \textit{Oracle}), we used a randomized sequence of augmentation steps (contrast shifts, flips) as a data augmentation strategy in the source domain. We did not use any augmentation for the target domain.

Quantitative evaluations and comparisons with state-of-the-art methods are reported. First, to evaluate the impact of domain shift on performance, we compared the proposed loss function to the baselines described in Section \ref{ssec:baselines}.  We then implemented our proposed CDA in a weakly supervised setting, as described in Section \ref{ssec:curi}. To juxtapose the performance of CDA to other domain adaptation methods under the same conditions, we provide quantitative and qualitative results of two current state-of-the-art methods for domain adaptation, using an adversarial  \cite{tsai2018learning} or a curriculum strategy \cite{zhang2017curriculum}.  We also report benchmark results from \cite{dou2018pnp} on MMWHS. Additionally, we provide a comprehensive analysis of the robustness of CDA to imprecision in the prior knowledge. In the first ablation study, we remove the size-regression network, and use source statistics as size priors instead. In the second, we remove the weak image-level tag annotation in the target domain, to test the robustness of our method in a fully unsupervised domain adaptation setting.

\subsubsection{\textbf{Baselines}}
\label{ssec:baselines}

\paragraph{\textbf{Lower and upper baselines}}To evaluate the impact of the different domain adaptation approaches, we trained a segmentation network in a fully-supervised manner, in both source and target images. Training these fully supervised models reduces to minimizing a standard loss function that evaluates the discrepancy between the CNN predictions and the corresponding ground-truth segmentations: $\min_{\theta}\sum_{d \in D} \sum_{i \in \Omega_d} {\cal L}({\mathbf y}_d (i), {\mathbf p}_d (i, \theta))$, where $D$ indicates the image domain. Thus, the model trained with the source images, i.e., $D$=$S$, will be referred to as \textit{NoAdap}, and will represent the lower baseline. The model employing the target images for training, i.e., $D$=$T$, will be denoted as \textit{Oracle}, and will serve as the upper baseline.

\paragraph{\textbf{Adversarial domain adaptation}}

We compared our CDA model to the adversarial approach proposed in \cite{tsai2018learning}, which has demonstrated state-of-the-art performances in the task of unsupervised domain adaptation for natural colour-image scenes. To do so, the penalty $\cal F$ in Eq. \eqref{eq:unconstrained-problem-penalty} is replaced by an adversarial loss, which enforces the alignment between the distributions of source and target segmentations. During training, non-aligned pairs of images from the source and target domain are fed into the segmentation network. Then, a discriminator uses the generated segmentation masks as inputs and attempts to identify the domain of each of these masks (either source or target).
In this setting, we focused on single-level adversarial learning (see \cite{tsai2018learning} for more details). 
For a fair comparison, we adopted and improved significantly the performance of the adversarial method in \cite{tsai2018learning}. 
Following the recent work in \cite{Pei2018,Chen2017}, in the weakly-supervised setting, we boosted the performance of \cite{tsai2018learning} using exactly the 
same image-level (tag) class information available to CDA. Specifically, we modified the discriminator loss so as to account for the tag of both source and target images. We experimented with various settings, and found that training the discriminator with only the positive images from both domains increased significantly the performance of \cite{tsai2018learning}. In fact, the use of negative (or mixed) pairs, in which the source and/or target images do not contain the region of interest, confuses adversarial training, reducing its performance in both applications.

For the adaptation of the cardiac sub-structure segmentation task, we also report the results by PnP-AdaNet \cite{dou2018pnp}, an adversarial method designed specifically for cross-modality DA composed of two independent domain-specific encoders and a decoder, along with a customized network architecture. \cite{dou2018pnp} also provided extensive comparisons to other state-of-the-art adversarial adaptation methods (DANN\cite{dann}, ADDA \cite{ADDA}, CycleGAN \cite{cyclegan}), used in conjunction with the same backbone segmentation architecture as in PnP-AdaNet, which we refer to as AdaNet. These methods are included in Table \ref{table:resultswhs} for comparison.

\paragraph{\textbf{Kullback-Leibler divergence adaptation}}
We further compare our approach to the recent curriculum domain method proposed in \cite{zhang2017curriculum}, which first learns region-proportion priors, i.e., label distributions. Then, the adaptation phase in \cite{zhang2017curriculum} is based on minimizing a Kullback–Leibler divergence, thereby matching the network predictions' label distributions to these priors. While conceptually related to our approach, the method in \cite{zhang2017curriculum} does not allow imprecision in the priors. Moreover, given the steeper profile of the Kullback–Leibler divergence compared to our penalty function in Eq. \eqref{eq:quadratic_penalty}, its optimisation may be less robust than CDA to a noisy prior. For a fair comparison, we used the same prior estimation obtained by the auxiliary regression network $R$ as in CDA (see Section \ref{ssec:curi}), and adapted the framework in \cite{zhang2017curriculum} to a weakly-supervised setting. Specifically, for each target image, the label distribution prior to be used for Kullback–Leibler divergence matching is:
 \begin{equation}
\hat{d_t}=\begin{cases}
    \frac{1}{\left|\Omega_{t}\right|} \hat{\tau_t} & \text{if region k is within image $t$}.\\
    0 & \text{otherwise}.
  \end{cases}
  \label{eq:curi}
\end{equation}
where $\hat{\tau_t}$ is the predicted size by $R$ on the target image $t$. For comparability and simplicity, we focused on matching label distributions at the image level, removing the additional superpixel level (see \cite{zhang2017curriculum} for more details).
In the experiments below, this method is referred to as \textit{KLAdap}.

\subsubsection{\textbf{Supervised Constraints}}
\label{sssec:supConst}

The proposed CDA method can accommodate both precise and imprecise (or uncertain) prior information about the target region, e.g. size, in the target domain. This is done by imposing inequality constraints of the general form in Eq. \eqref{eq:constrained-domain-adaption} on the target images. Such inequality constraints could be either tight, when we have precise priors, or loose otherwise. In all the following experiments, we imposed lower and upper bounds for each slice. We trained several models under the same setting, using different constraint values for the size priors on the target images.

First, we investigate the capability of the proposed CDA approach when precise information about the size of the segmentation regions is known. To this end, for each image $t$ and each structure $k$, of the target domain, we constrained the segmentation size by two prior values, which were derived from the ground-truth size $\tau_t$:
\begin{equation}
    \tau_t = \sum_{i \in \Omega_t} y_{t}^{1}(i)
\end{equation}
We start by introducing a relatively small uncertainty on this prior information, by adding a $\pm$ 10\% margin.
With the notations from Eqs. \eqref{constraint-upper-bound} and \eqref{constraint-lower-bound}, we have the lower and 
upper bounds:
\begin{equation}
a,b=\begin{cases}
    0.9\tau_t,1.1\tau_t & \text{if $\tau_t> 0$}.\\
    0,0 & \text{otherwise}.
  \end{cases}
\end{equation}
This setting is later on referred to as $Constraint_{10}$. Then, to evaluate the robustness of the proposed approach to imprecision in the prior knowledge about region size, we also investigated the effect of different levels of tightness of the bounds, by allowing larger margins from the exact size. In particular, we trained additional models with margins on the bounds equal to $\pm$ 25\%, $\pm$ 50\%, and $\pm$ 75\%, which are referred to as $Constraint_{25}$, $Constraint_{50}$ and $Constraint_{75}$, respectively. The aim of this setting is to evaluate how precise the target size information should be. This is different from the main experiments, where the ground truth target size $\tau_t$ is unknown.

\subsubsection{\textbf{Learning constraints via an auxiliary task}}\label{ssec:curi}
Instead of using bounds derived from the ground-truth size, in this setting, we employ bounds derived from the size estimations produced by the regression model $R$ introduced in Section \ref{sec:constraints}. In addition to the fully-labeled masks for the source images, we assume that weak image-level annotations are available for the target-domain images. These are image tags indicating whether a given image $t$ contains a region of interest $k$ or not. 
For each target image, the bounds to be used for adapting the segmentation network with constraints \eqref{constraint-upper-bound} and \eqref{constraint-lower-bound} are:
 \begin{equation}
a,b=\begin{cases}
    0.9\hat{\tau_t},1.1\hat{\tau_t} & \text{if region k is within image $t$}.\\
    0,0 & \text{otherwise}.
  \end{cases}
  \label{eq:curi_constraint}
\end{equation}
where $\hat{\tau_t}$ is the predicted size by $R$ on the target image $t$. This setting will be referred to as \textit{ConstraintAdap} in the following experiments.

\subsubsection{\textbf{Deriving constraints from estimated anatomical knowledge}}\label{ssec:noregressor}
To investigate to possibility to circumvent the auxiliary network, in this setting, we derive the size estimations from the source statistics, which aims at approaching textbook anatomical knowledge. For each 2D target image $t$ and each structure $k$, the bounds to be used for adapting the segmentation network with constraints \eqref{constraint-upper-bound} and \eqref{constraint-lower-bound} are:
 \begin{equation}
a,b=\begin{cases}
    0.9\bar{\tau}_S,1.1\bar{\tau}_S & \text{if region $k$ is within image $t$}.\\
    0,0 & \text{otherwise},
  \end{cases}
  \label{eq:constraint_lit}
\end{equation}
where $\bar{\tau}_S$ is the median of ground truth sizes for structure $k$ in the source images $I_s$ from the training set. We refer to this ablation study as \textit{ConstraintLit} in the following.

\subsubsection{\textbf{Evaluation on Segmentation Performance}}
In all our experiments, we employed two commonly-used metrics to quantitatively evaluate the segmentation performance of models. First, the Dice similarity coefficient (DSC) which evaluates the degree of overlap between the segmentation regions and the ground truth. Second, the Hausdorff distance (HD) which measures boundary distances. We used the 95-th percentile of the Hausdorff distance (HD95) to mitigate noisy ground truth and/or segmentation regions. Therefore, higher DSC values, and lower HD95 values indicate better segmentation performances. As the data is volumetric, these metrics were computed over the 3D segmentations.

\subsubsection{\textbf{Training and Implementation Details}}\label{ssec:trainingdetails}

For the segmentation networks, we employed ENet \cite{paszke2016enet}, since it achieves good segmentation performance in a reduced time. We also showed additional results with UNet in order to compare with a different backbone architecture. We employed the standard cross-entropy (CE) for the source segmentation loss, along with results combined with the DiceLoss ($Dice+CE$) \cite{Sudre_2017,Milletari}. All adaptation models were initialized by training the network with the segmentation loss only, on the source domain, for 150 epochs. For $\gamma$ in Eq. \eqref{eq:unconstrained-problem-penalty} we adopt a grid search to choose the best value of the weighting $\gamma$ parameter for each setting. In the adversarial DA setting, we employ the same segmentation network and include the discriminator proposed in \cite{tsai2018learning}, while for \textit{KLAdap}, the Kullback-Leibler divergence in \cite{zhang2017curriculum} was included. In all the domain adaptation experiments, we use Adam optimizer for minimizing the respective loss functions, with an initial learning rate of $1\times 10^{-3}$. The best model was chosen based on the validation set.

Finally, in the \textit{ConstraintAdap} and \textit{KLAdap} settings, the regression network used to learn the size prior is a ResNeXt 101 \cite{resnext},    trained from scratch. We trained it via standard stochastic gradient descent, with a learning rate of $5\times 10^{-6}$. The code is implemented in PyTorch. We ran the experiments on a machine equipped with an AMD Ryzen 1950X 16-Core Processor, 32 GB of RAM and an NVIDIA Titan XP GPU.

\subsection{Results}

\paragraph*{\textbf{Quantitative results}} \label{sec:quantires}

Table \ref{table:resultsivd} reports the quantitative performance of different methods in spine images. With ENet as the backbone architecture, we observe that \textit{NoAdap} achieves the worst performance, with a 46.8\% mean DSC. This is not surprising, since the distributions of source and target images are significantly different due to the presence of the domain shift. This indicates that direct transfer of segmentation models trained on the source cannot handle properly the domain gap.  
Adopting an adversarial strategy allows to stabilize and improve the results over the lower baseline, achieving a mean DSC of 57.3\%.
The largest improvement is observed when the domain adaptation strategy incorporates a constrained term on the target predictions. First, we observe in Table \ref{table:results_supcons} that if the target size is known, the DSC obtained by \textit{Constraint}$_{10}$ is 80.4\%, which corresponds to 95\% of the full supervised model, i.e., \textit{Oracle}. However, knowing with precision the size of the structure to be segmented is not always feasible. In the more realistic scenario \textit{ConstraintAdap}, where this size prior is estimated, the mean DSC value only drops to 72.3\%, achieving 86\% of the performance of the upper bound, \textit{Oracle}. Moreover, our model outperforms \textit{KLAdap} by 3.5\%, which uses the same size prior estimation but a Kullback-Leibler divergence as a regularisation loss. This demonstrates the usefulness of using inequality constraints around the estimated prior and a less aggressive loss such as Eq. \ref{eq:quadratic_penalty}.
The HD values present a similar pattern across the different models. While the adversarial approach reduced the HD to the half (10.3 mm) compared to the lower baseline model (20.7 mm), \textit{KLAdap} obtained a HD of 6.3 mm. Our proposed model \textit{ConstraintAdap} further improved the results, achieving a HD of 5.4 mm.
To demonstrate that our approach is model-agnostic, and generalizes well to other architectures, we replace ENet by UNet. We observe that the results are consistent with those observed with ENet. In particular, while replacing the segmentation network by UNet brings performance gains across methods, the rankings are maintained, with our approach outperforming prior state-of-the-art models. Last, even though the improvement gap is lesser with UNet, it still shows the effect of domain transfer across the various training settings.

Table \ref{table:resultswhs} presents the results for segmentation in cardiac images from the MM-WHS MICCAI Dataset. With no adaptation strategy, the performance of a model learnt on MRI images degrades when it is tested on CT images, with an average DSC of 17.3\%. 
On the other hand, our proposed \textit{ConstraintAdap} achieves an average HD of 11.0 mm, and an average DSC of 71.4\%, representing 80\% of the upper bound model, i.e., $Oracle$, trained on target images. Our method significantly outperforms other state-of-the-art approaches on both metrics. Specifically, the adversarial method in \cite{tsai2018learning} only yields a 41.1\% average DSC and an average HD of 45.9 mm, whereas \textit{KLAdap} obtains closer values, with an average DSC of 70.7\%, and an average HD of 12.8 mm. Quantitative results from \cite{dou2018pnp} are also provided, which shows that our method outperforms prior works also on this dataset.
It is important to highlight that direct comparison is not appropriate, as the backbone architecture of these methods is different.

\begin{figure}[t]
 \begin{center}
 \mbox{
 \includegraphics[width=0.8\linewidth]{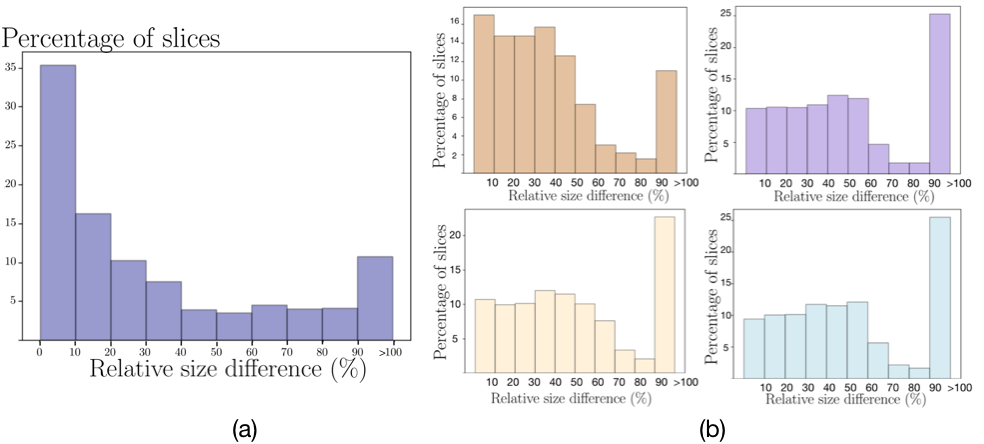}
 }
\caption{Normalized histograms of the relative size difference between ground truth size and size estimated by the auxiliary task in the target domain for spine images (a) and cardiac images (b, clockwise for Myo, LA, AA, LV). This size estimation is used as a prior to guide domain adaptation (see Section \ref{ssec:curi}).} 
\label{fig:sizehist}
\end{center}
\end{figure}

Finally, we should note that in both applications, the size estimation obtained by the size regression network $R$ is quite noisy, as shown in Figure \ref{fig:sizehist}. This suggests robustness to prior imprecision of CDA models as we further explore in the ablation study below.

\begin{table}[t]
\centering
\footnotesize
\caption{Performance comparison of the proposed formulation with different domain adaptation methods for spine segmentation.}
  \vspace{3mm}
\setlength\tabcolsep{0.5pt} 
\begin{tabular}{llcccc}
\toprule
\textbf{Backbone}&   \textbf{Methods} &\textbf{Target}& \textbf{DSC(\%) } &\textbf{HD95(mm) }\\
\multirow[-3pt]{1}{*} 
&   &\textbf{Tags} & \textbf{mean$\pm$sd}  & \textbf{mean$\pm$sd}    \\
\midrule \midrule

 \multirow{5}{*}{\begin{tabular}[c]{@{}l@{}}ENet\end{tabular}}&
 NoAdap & $\times$  & 46.8$\pm$11.1  & 20.7$\pm$6.7 \\
&  Adversarial\cite{tsai2018learning} &\checkmark &  57.3$\pm$6.5 & 10.3$\pm$2.6 \\
&  KLAdap\cite{zhang2017curriculum}  &\checkmark&   68.8$\pm$2.2 & 6.3$\pm$0.5  \\
 &ConstraintAdap (ours)&\checkmark& \textbf{72.3$\pm$2.6}  & \textbf{5.4$\pm$1.5} \\
 &  Oracle &\checkmark& 84.5$\pm$1.6 & 3.0$\pm$0.3   \\
\midrule
\multirow{5}{*}{\begin{tabular}[c]{@{}l@{}}UNet\end{tabular}} &
NoAdap &$\times$& 63.9$\pm$7.5  & 9.6$\pm$7.0  \\
&  Adversarial\cite{tsai2018learning} &\checkmark &  69.0$\pm$3.8 & 6.4$\pm$1.5\\
&  KLAdap\cite{zhang2017curriculum} &\checkmark &  73.3$\pm$1.5 &  5.9$\pm$2.0 \\
&ConstraintAdap (ours)&\checkmark & \textbf{73.4$\pm$2.4} & \textbf{4.7$\pm$0.9}  \\
 &  Oracle &\checkmark& 85.4$\pm$3.0  & 2.3$\pm$0.3 \\
\bottomrule\\[6pt]
\end{tabular}

\label{table:resultsivd}
\end{table}

\begin{table*}[t]

\footnotesize
  \caption{Performance comparison of the proposed formulation with different domain adaptation methods for cardiac segmentation, in terms of DSC (mean$\pm$std) and HD (mean$\pm$std).  (Note: - means that the results are not reported in the original papers)}
   \begin{adjustwidth}{-1cm}{-1cm}
   \centering
  \vspace{3mm}
  \setlength\tabcolsep{0.1pt}
  \begin{tabular}{llccc|cc|cc|cc|cc}

\toprule

\textbf{Backbone}& \textbf{Methods} &  \textbf{Target}& \multicolumn{2}{c}{\textbf{Myo}} & \multicolumn{2}{c}{\textbf{LA}} & \multicolumn{2}{c}{\textbf{LV}} & \multicolumn{2}{c}{\textbf{AA}} & \multicolumn{2}{c}{\textbf{Mean}}
\\

& & \textbf{Tags}&\textbf{DSC} & \textbf{HD}& \textbf{DSC} & \textbf{HD}& \textbf{DSC} & \textbf{HD}& \textbf{DSC} & \textbf{HD}& \textbf{DSC} & \textbf{HD} \\
\midrule \midrule
\multirow{5}{*}{\begin{tabular}[c]{@{}l@{}}ENet\end{tabular}} 
& NoAdap & $\times$ & 9.0$\pm$11.2 & 38.6$\pm$13.5 & 26.2$\pm$30.2 &  60.5$\pm$16.3 & 1.5$\pm$2.3 & 36.2$\pm$24.0 & 32.6$\pm$37.2 & 62.8$\pm$14.1 & 17.3$\pm$20.2 & 49.5$\pm$17.0\\
& Adversarial \cite{tsai2018learning} & \checkmark & 19.3$\pm$8.3 & 32.5$\pm$10.9 & 58.1$\pm$14.5 & 54.8$\pm$22.4  & 22.5$\pm$18.6 & 45.9$\pm$15.8 & 64.5$\pm$14.8 & 50.4$\pm$8.7 & 41.1$\pm$14.0& 45.9$\pm$14.5\\ 
&  KLAdap \cite{zhang2017curriculum}&\checkmark& 61.3$\pm$6.5 & \textbf{11.2$\pm$5.1}  & \textbf{77.9$\pm$7.3}  & 21.6$\pm$24.1 & 64.4$\pm$10.7& \textbf{9.8$\pm$2.4} & \textbf{79.0$\pm$5.6} & \textbf{8.5$\pm$0.9} & 70.7$\pm$7.6 & 12.8$\pm$8.1 \\
& ConstraintAdap(ours)& \checkmark & \textbf{61.9$\pm$0.9} & 16.1$\pm$6.5 & 72.8$\pm$12.6 & \textbf{8.7$\pm$2.7} & \textbf{73.7$\pm$5.7}& 10.0$\pm$3.1 & 77.3$\pm$6.2& 9.0$\pm$1.9  & \textbf{71.4$\pm$6.4} & \textbf{11.0$\pm$3.5} \\ 
 &  Oracle  & \checkmark & 85.8$\pm$ 3.3& 2.9$\pm$1.6 & 90.1$\pm$3.2 & 4.5$\pm$3.0 & 90.8$\pm$3.4&  3.0$\pm$1.5 & 91.0$\pm$7.3 & 2.9$\pm$1.6 & 89.4$\pm$4.3 & 3.3$\pm$1.9\\
 \midrule
\multirow{5}{*}{\begin{tabular}[c]{@{}l@{}}AdaNet\end{tabular}} 
&  NoAdap  & $\times$&   15.3$\pm$17.2 & - & 2.7$\pm$0.8 & - & 3.4$\pm$5.8 &- & 31.5$\pm$23.9 & - & 13.2$\pm$11.9 &- \\
&   DANN \cite{dann} & $\times$&     25.7$\pm$13.2 & -& 45.1$\pm$23.6 & -& 28.3$\pm$11.8 & -& 39.0$\pm$35.1 & -& 34.5$\pm$20.9 &-\\
&  ADDA \cite{ADDA} & $\times$&   29.2$\pm$16.4 & -&60.9$\pm$13.2& -& 11.2$\pm$13.1& -& 47.6$\pm$15.2 & -& 37.2$\pm$14.5&- \\
&  CycleGAN \cite{cyclegan}  & $\times$&   28.7$\pm$13.3& -& \textbf{75.7$\pm$ 4.3}  & -& 52.3$\pm$21.0& -&  73.8$\pm$7.4 & -&57.6$\pm$11.5 &-\\
&  PnP-AdaNet \cite{dou2018pnp}  & $\times$&   \textbf{50.8$\pm $7.0}& -& 68.9$\pm$ 5.2& -& \textbf{61.9$\pm$10.7}& -& \textbf{74.0$\pm$7.3} & -& \textbf{63.9$ \pm$ 7.5}&- \\
\bottomrule\\[6pt]
  \end{tabular}
  \label{table:resultswhs}
   \end{adjustwidth}
\end{table*}

\begin{table}[t]
\centering
\footnotesize
\caption{Performance comparison for the proposed formulation with constraints derived from the ground truth (Constraint$_{25,50,75}$) and from the source-domain statistics (Constraint$_{Lit}$). ENet is employed as backbone architecture.}
\vspace{3mm}
\setlength\tabcolsep{5.5pt} 
\begin{tabular}{llccc}
\toprule
\textbf{Dataset} &\textbf{Methods} &\textbf{DSC (\%)} &\textbf{HD95(mm)}  \\
\multirow[-3pt]{1}{*} 
&     & \textbf{mean$\pm$sd} & \textbf{mean$\pm$sd} \\
\midrule 
\midrule
\multirow{7}{*}{\begin{tabular}[c]{@{}l@{}}IVDM3Seg\end{tabular}}   & NoAdap & 46.8$\pm$11.1 &  20.7$\pm$6.7  \\
 & Constraint$_{Lit}$  & 60.7$\pm$2.8  & 7.2$\pm$2.4 \\
 & Constraint$_{75}$  & 65.7$\pm$4.2  & 6.9$\pm$1.7 \\
 & Constraint$_{50}$  & 74.6$\pm$2.1  & 4.5$\pm$0.7  \\
 & Constraint$_{25}$  & 77.5$\pm$0.7  & 4.1$\pm$0.5 \\
 & Constraint$_{10}$ & \textbf{80.4$\pm$1.5} &  \textbf{3.7$\pm$0.9}\\
 & Oracle  & 84.5$\pm$1.6 & 3.5$\pm$0.3 \\
\midrule
\multirow{7}{*}{\begin{tabular}[c]{@{}l@{}}MMWHS\end{tabular}}   & NoAdap & 38.2$\pm$11.4 &N/A$^{\mathrm{a}}$ \\ 
&Constraint$_{Lit}$  & 64.2$\pm$6.0  &11.6$\pm$6.4 \\
&Constraint$_{75}$  & 69.4$\pm$11.1  &10.4$\pm$5.9  \\
&Constraint$_{50}$  & 79.9$\pm$6.8  & 7.5$\pm$2.6  \\
&Constraint$_{25}$  &  82.5$\pm$6.4 & 7.5$\pm$5.3  \\
& Constraint$_{10}$ & \textbf{84.6$\pm$5.3} &  \textbf{7.2$\pm$6.0} \\
& Oracle  & 89.4$\pm$4.3&6.5$\pm$5.6 \\
\bottomrule\\[-6pt]
\multicolumn{4}{p{200pt}}{$^{\mathrm{a}}$N/A means that the value cannot be calculated due to no prediction for that structure.}
\end{tabular}
\label{table:results_supcons}
\end{table}

\begin{figure*}[t]
\centering
\includegraphics[width=1\linewidth]{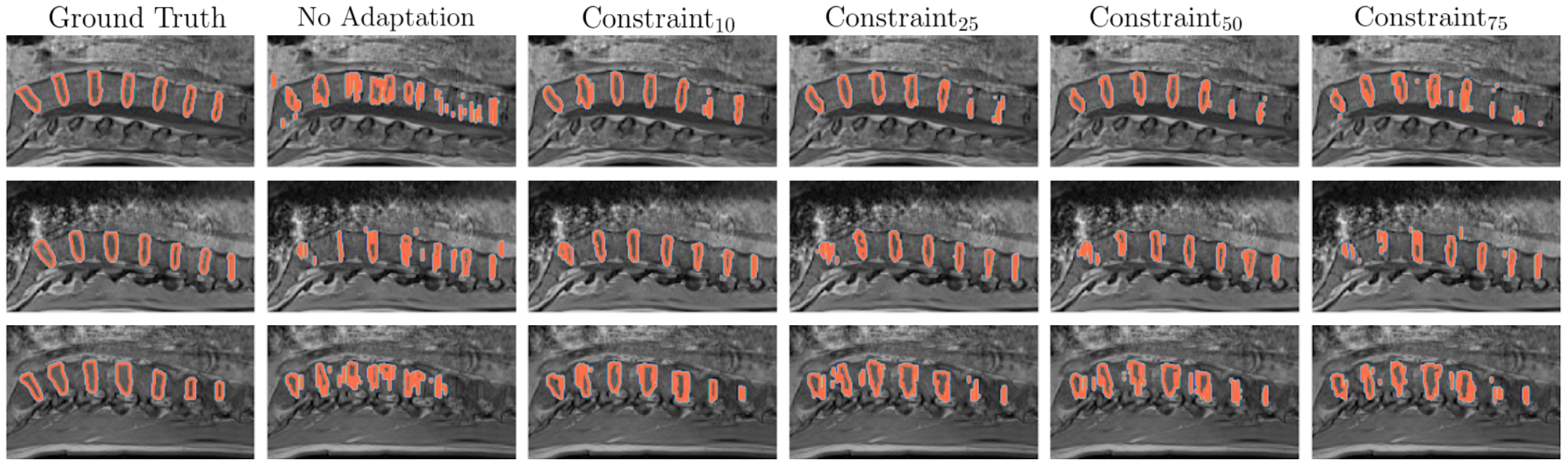}
\vspace{-7mm}
\caption{Example of the segmentations achieved on spine images by our constrained formulation with tighter to looser constraints  (\textit{Constraint}$_{10}$ being the tightest), i.e., increasing prior uncertainty, showing robustness to prior imprecision. Each row shows a different test subject. Images and masks are rotated in the sagittal plane and cropped for better viewing. The IVDs are contoured in red.}
\label{fig:segmmargins}
\end{figure*}

\begin{figure*}[t]
\centering
\includegraphics[width=0.8\linewidth]{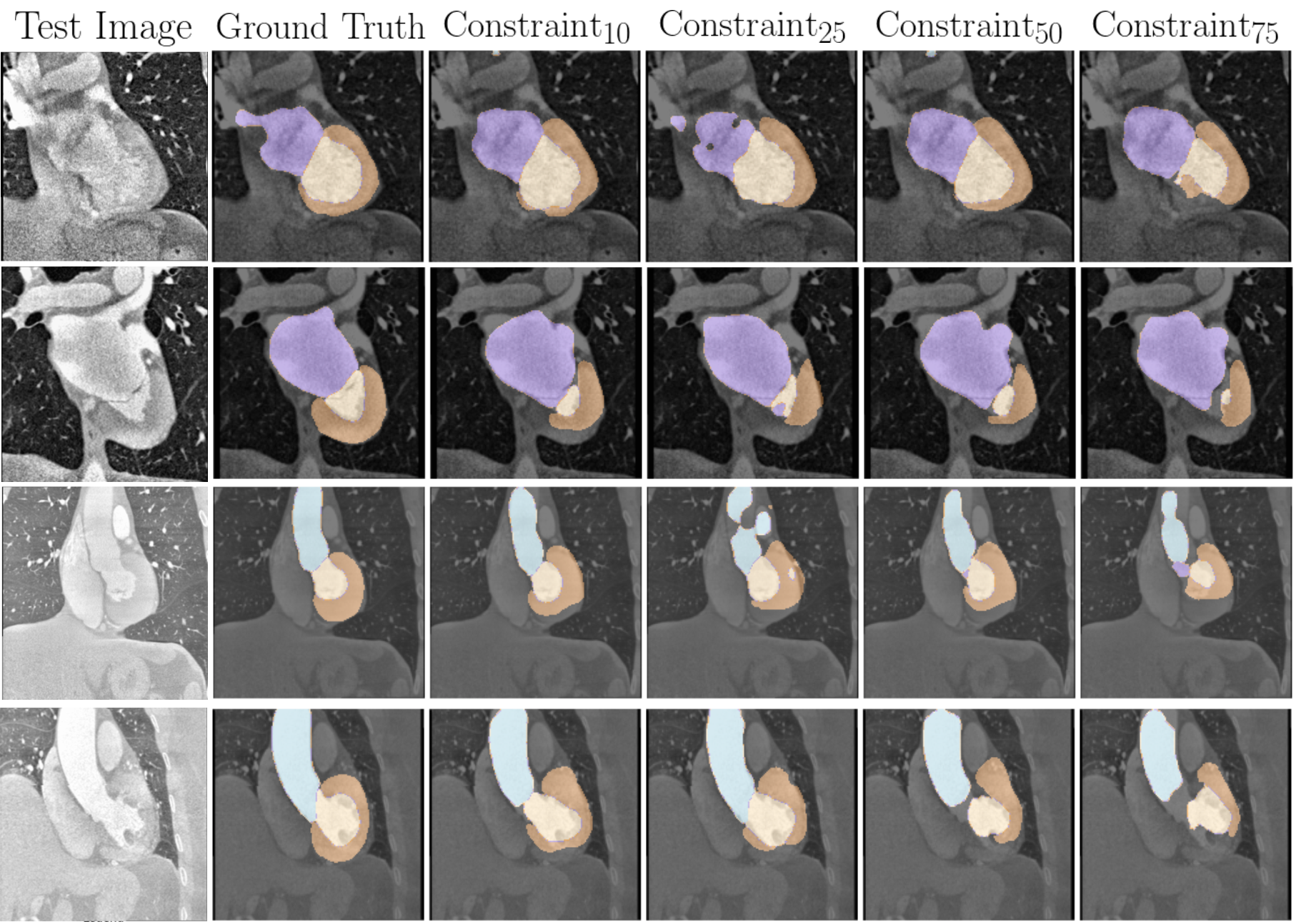}
\vspace{-3mm}
\caption{Examples of the segmentations achieved on cardiac CT images by our constrained formulation with tighter to looser constraints (\textit{Constraint}$_{10}$ being the tightest), i.e., increasing prior uncertainty, showing robustness to prior imprecision.}
\label{fig:segmmarginswhs}
\end{figure*}

\paragraph{\textbf{Ablation study on bound precision}}
\label{sssec:abla}
We also investigated the impact of prior size imprecision in the target domain on the quality of CDA models. To this end, we increase the lower and upper margins around the true size, as explained in Section \ref{sssec:supConst}. Results from this study are reported in Table \ref{table:results_supcons} and in Figures \ref{fig:segmmargins} and \ref{fig:segmmarginswhs}. As expected, having precise size constraints result in higher performing models, close to the full-supervision setting on target images. Nevertheless, allowing large ambiguities on the size of the region of interest ($\pm$ 25-50\%) only degrades the DSC performance by up to 6\% on spine images, and to 5\% on cardiac images.
In the ablation study $ConstraintLit$, we replaced the size regressor by simple source statistics as explained in Section \ref{ssec:noregressor}. Interestingly, results are well above the baseline for spine and cardiac images, yielding 60.7\% DSC and 64.2\% average DSC, respectively. This indicates that having a coarse knowledge of the target size can be enough to guide adaptation with CDA. Furthermore, if the target image tag is available, it is possible  to circumvent the auxiliary network size regressor $R$.

\begin{table}[t]
\centering
\footnotesize
\caption{Performance of the different domain adaptation methods obtained when removing the weak image-tag annotations. ENet is employed as backbone architecture.}
\vspace{3mm}
\setlength\tabcolsep{5pt} 
\begin{tabular}{llcccccc}
\toprule
\textbf{Dataset} & \textbf{Methods} &\textbf{DSC (\%)}  & \textbf{HD95 (mm)}  \\
\multirow[-3pt]{1}{*} 
&   & \textbf{mean$\pm$sd}  & \textbf{mean$\pm$sd}   \\
\midrule \midrule

\multirow{3}{*}{\begin{tabular}[c]{@{}l@{}}IVDM3Seg\end{tabular}} 
 &  Adversarial\cite{tsai2018learning}  &  48.7$\pm$2.4 & 18.0$\pm$7.8 \\
 &  KLAdap\cite{zhang2017curriculum}  &   52.2$\pm$5.9 & 12.5$\pm$4.2   \\
 & ConstraintAdap(ours)  & \textbf{58.3$\pm$2.1}  & \textbf{7.4$\pm$3.1} \\
\midrule
\multirow{3}{*}{\begin{tabular}[c]{@{}l@{}}MMWHS\end{tabular}} 
& Adversarial\cite{tsai2018learning} &   38.9$\pm$14.6 &  \textbf{32.1$\pm$10.4}   \\
& KLAdap\cite{zhang2017curriculum}  &    33.3$\pm$13.8 & N/A   \\
& ConstraintAdap(ours)   & \textbf{49.4$\pm$14.8}  & 42.3$\pm$11.5  \\
 \bottomrule\\[6pt]
\end{tabular}
\label{table:resultsnotag}
\end{table}

\begin{table}[t]
\centering
\footnotesize

\caption{Performance comparison of the proposed formulation with different segmentation losses defined over the source spine data. UNet is employed as backbone architecture.}
  \vspace{3mm}
\setlength\tabcolsep{6pt} 
\begin{tabular}{lccccc}
\toprule
 \textbf{Method} &\textbf{Source}&  \textbf{DSC(\%) } &\textbf{HD95(mm) }\\
\multirow[-3pt]{1}{*} 
 & \textbf{Loss}&  \textbf{mean$\pm$sd}  & \textbf{mean$\pm$sd}    \\
\midrule 
NoAdap &CE& 63.9$\pm$7.5  & 9.6$\pm$7.0 \\
ConstraintAdap (ours)&CE& 73.4$\pm$2.4& 4.7$\pm$0.9  \\
\midrule
NoAdap &Dice+CE& 65.5$\pm$2.1  & 6.0$\pm$0.8\\
ConstraintAdap (ours)&Dice+CE& \textbf{75.7$\pm$1.8} & \textbf{4.7$\pm$0.7}  \\
\bottomrule\\[6pt]
\end{tabular}

\label{table:resultsdiceloss}
\end{table}

\paragraph{\textbf{Assessing the impact of the target image tags}}\label{sssec:notag}

We investigated the effect of removing the image-level tag annotation in the target domain. Particularly, we removed the target image tags for both the size regressor and the adaptation phase, as explained in Section \ref{ssec:curi}. Results from this study are reported  in Table \ref{table:resultsnotag}. As expected, having image-level tag information considerably helps all the models, which can be observed from the performance decrease in comparison to the results in Table \ref{table:resultsivd} and \ref{table:resultswhs}. Indeed, the size estimation degrades without the image tag and, as a result, models using a size prior to guide adaptation also see their performance decrease.

An interesting observation in this scenario, however, is the larger gap between the proposed model and prior work, particularly compared to \textit{KLAdap}.

\paragraph*{\textbf{Qualitative results}}

\begin{figure*}[t]
\centering
\includegraphics[width=1\linewidth]{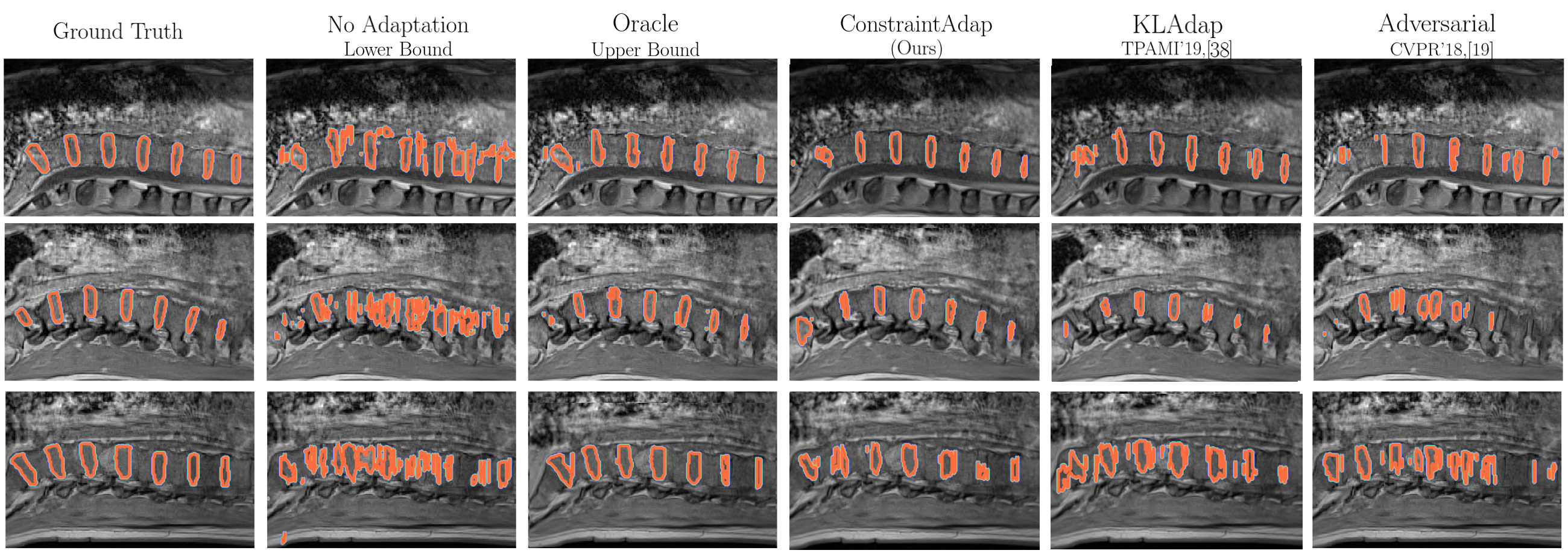}
\vspace{-7mm}
\caption{Example of the segmentations achieved by our constrained formulation (\textit{ConstraintAdap}), benchmark models in \cite{zhang2017curriculum} and \cite{tsai2018learning} and lower (\textit{NoAdap}) and upper baselines (\textit{Oracle}) for intervertebral disks images in the MRI In-Phase modality. Each row shows a different test subject. Images and masks are rotated in the sagittal plane and cropped for better viewing. The IVDs are contoured in red.}
\label{fig:segmivd}
\end{figure*}

\begin{figure*}[t]
\centering
\includegraphics[width=1\linewidth]{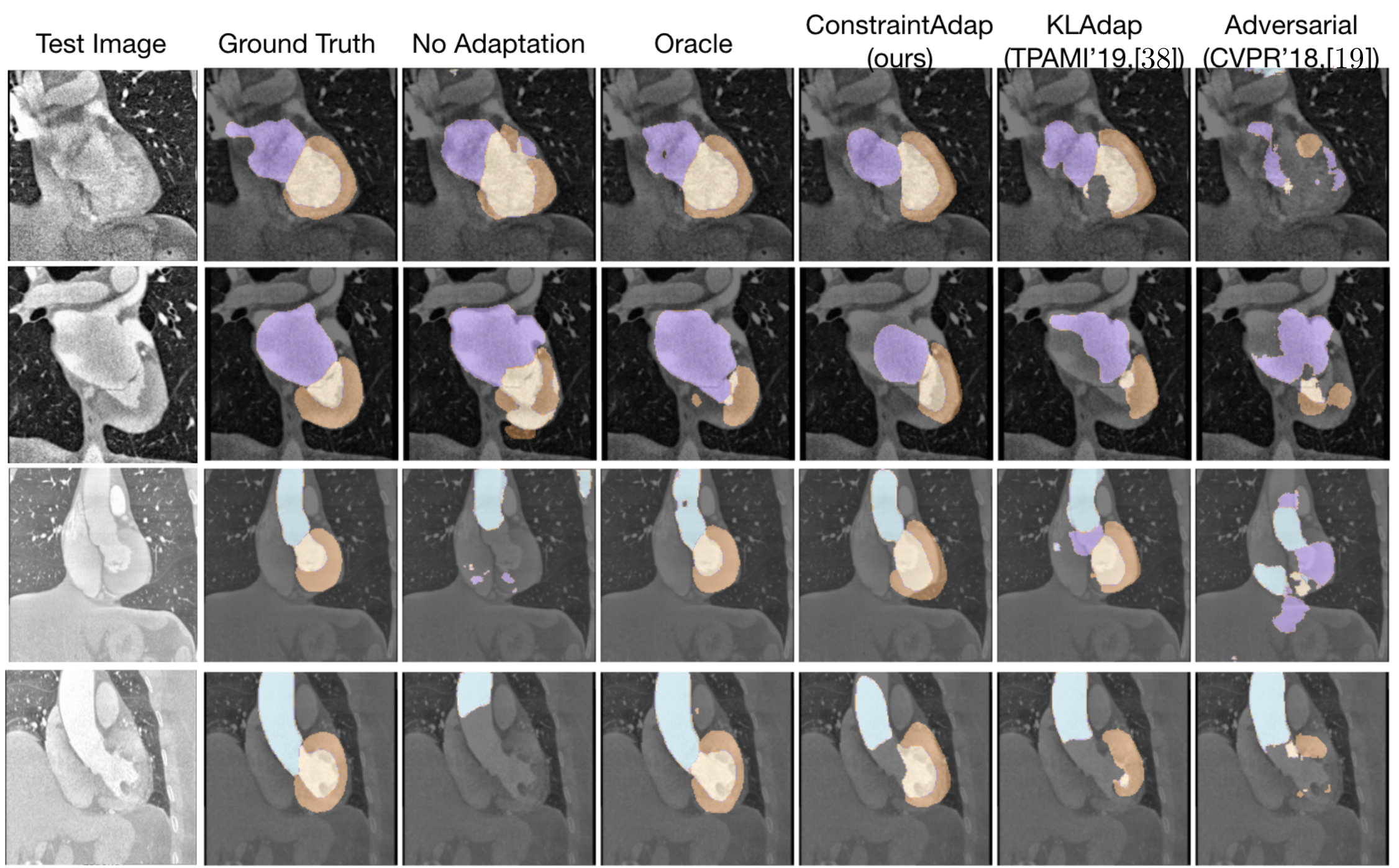}
\vspace{-7mm}
\caption{Examples of the segmentations achieved by our constrained formulation (\textit{ConstraintAdap}), benchmark models in \cite{zhang2017curriculum} and \cite{tsai2018learning} and lower (\textit{NoAdap}) and upper baselines (\textit{Oracle}) for cardiac CT images. The cardiac structures of MYO, LA, LV and AA are depicted in brown, purple, yellow and blue, respectively. Each row shows a different test subject.}
\label{fig:segmwhs}
\end{figure*}

Figure \ref{fig:segmivd} and \ref{fig:segmmargins} depict visual segmentation results for spine images, for the 3 subjects used in validation sets. We visualize the results at the best epoch. It can be seen that without adaptation, the network trained only on source images is unable to recover the 7 distinct IVDs present in all the subjects, and the model trained with adversarial adaptation also struggles (see the second row in Figure \ref{fig:segmivd}). In contrast, our proposed CDA model - both with supervised and learned constraints - is able to detect the 7 IVD structures in almost all examples. Moreover, the segmentations achieved by all models using the proposed CDA framework have more regular shapes. 
Figure \ref{fig:segmwhs} and \ref{fig:segmmarginswhs} show the visual comparison results on the cardiac dataset, for the 4 subjects in the test set. As illustrated in Figure \ref{fig:segmwhs}, the segmentation results produced by \textit{ConstraintAdap} are more similar to the ground truth, in terms of shape and boundary, especially for the MYO and LV structures.
Finally, we can visually observe in Figure \ref{fig:segmmargins} and \ref{fig:segmmarginswhs} that all constrained models $Constraint_{10}$, ..., $Constraint_{75}$, yield much better segmentations than the lower baseline without any adaptation strategy. Furthermore, as expected, the quality of the segmentations slowly degrades with a more imprecise size prior used for constraining the adaptation.

\paragraph*{\textbf{Efficiency}}
The computational efficiency of constrained formulations, benchmark adaptation formulations and baselines are compared in Table \ref{table:time}. The lower (\textit{NoAdap}) and upper (\textit{Oracle}) baselines only need to compute one loss per pass, i.e., cross-entropy, and only use images from the domain on which it is calculated, i.e., source (\textit{NoAdap}) or target (\textit{Oracle}), respectively. As expected, training times are lower for these methods. All other methods employ images from both domains at each forward pass. Including the quadratic loss with supervised size constraints adds little to the computational time. Using learned priors, such as in models \textit{ConstraintAdap} and \textit{KLAdap}, does not significantly change the computational time either, even when including the size-regressor training. Particularly, if we consider a two-step process, assuming the same number of epochs for all the models, the proposed constrained framework is still nearly twice faster than the adversarial approach in \cite{tsai2018learning}.
The overhead is much higher with the adversarial adaptation, which alternates at each pass between the training of the segmentation network and the training of the discriminator, the latter also requiring inputs from both domains. 

\begin{table}[]
\centering
\footnotesize
\caption{Training times  of the various adaptation learning strategies and \textit{Oracle} for a batch size of 12, for spine segmentation}
\vspace{3mm}
\begin{tabular}{llcc}
\toprule
\textbf{Backbone} &\textbf{Methods}& \textbf{Average Time (s/batch)}  \\
\midrule
ResNeXt101& $R$ (size regressor) &  0.2 \\
\midrule
\multirow{6}{*}{\begin{tabular}[c]{@{}l@{}}ENet\end{tabular}} & NoAdap   & 0.4   \\
& Adversarial\cite{tsai2018learning}   &1.4  \\
& KLAdap\cite{zhang2017curriculum}  &0.6   \\
& ConstraintAdap (ours)  &0.6    \\
& Constraint$_{10,25,50,75}$  &0.6   \\
& Oracle   &0.4 \\
\bottomrule\\[-18pt]
\end{tabular}
\label{table:time}
\end{table}

\section{Discussion}
\label{sec:discussion}

We presented a method to guide a segmentation network learned on a source domain to perform well on a different target domain, with minimal additional information, for instance, in the form of image-level tags. We showed the versatility of our DA approach, implementing it for drastically different types of images, multi-modal spine MRI images and MRI to CT cardiac images. Our model consistently yields a performance gain of 1-4\% in terms of DSC across architectures and datasets, and 4-14\% when comparing to state-of-the-art adversarial adaptation approaches.
Even though we have evaluated our method on multi-modal (multi-MRI and CT to MRI) spine and cardiac images, it can be applied to other multi-modal scenarios, such as multimodal photoacoustic and optical coherence tomography \cite{Hojjatoleslami}, for example. Unlike adversarial strategies, which are based on two-step training, our method tackles the adaptation problem with a single constrained loss, simplifying the adaptation of the segmentation network. In our implementation, the constrained loss matches image-level statistics--the size of the structure to be segmented here--in the target domain through the use of a simple quadratic loss. As demonstrated in our experiments, the performance is significantly improved over the lower baseline. Surprisingly, state-of-the-art adversarial methods \cite{tsai2018learning,ADDA,dann,cyclegan} yield smaller improvements. We hypothesize that this is due, in part, to the difficulty of learning a decision boundary between source and target domains in huge dimensionality. When a very precise size prior is known on the target domain, our framework leveraged this information to improve results up to 95\% of the upper bound (the full supervision regime on the target) on two different tasks. As shown quantitatively and qualitatively by our experiments, the structures of interests are much better detected in each patient in the target domain, while the segmentations achieved are greatly improved. Furthermore, we have shown that our method tolerates a substantial imprecision around the true size of structures, and that we can learn a sufficiently accurate size prior with a simple regression network. Although the estimated size prior obtained in our application is quite noisy, and our uncertainty margins very simple, our formulation with learned constraints reaches 86\% and 80\% of full supervision in spine and cardiac images respectively. We also demonstrate the superiority of our method compared to a domain adaptation model using size statistics matching with a steeper loss and no handling of prior imprecision \cite{zhang2017curriculum}.

Arguably, the main limitation of our method relies on obtaining an accurate estimation of region size, which guides segmentation training during the phase of domain adaptation. Learning region size through an auxiliary regression network could be challenging when there is a large shift between the source and target domains. However, we show in Table \ref{table:results_supcons} that, even with large ambiguities on size estimation, the performance of the proposed model drops by only 5-6\% on both datasets. An interesting finding from our results is that adding the weak image-level class  information, i.e.,  the  presence  or  absence of the target region, for each slice in the target domain greatly helped the auxiliary size regressor network. 
The need of this weak annotation to approach the performances of full supervision might be seen as another drawback of our method.
This contrasts with fully unsupervised domain adaptation methods, which do not require weak annotations, but are usually unstable and hard to train. Nevertheless, we showed in Table \ref{table:resultsnotag} that a fully unsupervised version of our method, without access to image-tag information, still outperforms several state-of-the-art adaptation methods based on adversarial training \cite{zhang2017curriculum,tsai2018learning}. We argue that, despite these drawbacks, our method provides an optimization framework that is simpler and more stable than fully unsupervised adaptation methods.

Future developments could involve a 3D extension, for which some questions related to the incorporation of textbook medical knowledge remain undefined, as it is common to use volumes patches as input to 3D networks \cite{DolzNeuro2017}.  Learning other priors from domain information, such as constraints derived from shape moments \cite{Klodt}, and better addressing the uncertainty of size estimations, for instance, from deriving more sophisticated margins, are other potential improvements left for future work. For difficult domain adaptation tasks with substantial domain shift, the initial network trained on the source domain may be incapable of detecting any structure in the target domain, complicating the initialization of our method. In such cases, as an alternative approach, weak annotations such as bounding boxes \cite{Rajchl2016,Dai2015} in the target domain could be used. Another open question for such difficult applications is the usefulness of enforcing multiple constraints and how to handle the ensuing optimisation problem.

\section{Conclusion}

This study investigated domain adaptation for segmentation with applications for intervertebral discs segmentation in multi-modal MRI and MRI to CT cardiac substructure segmentation. We proposed a constrained formulation for adapting a segmentation network learned on one modality (source domain) to a different modality (target domain), by enforcing image-level statistics in the target domain which we showed could be learned directly from the source domain. Despite its simplicity, the performance of our method comes near that of full supervision with only image-level annotations in the target domain, and very small computation overhead, using basic linear constraints, e.g., target-region size. Extensive experiments demonstrated that our formulation also outperformed multiple state-of-the-art adaptation methods.
Our framework offers, therefore, flexibility, is model-agnostic and opens the door to promising research directions on incorporating a wide variety of new anatomical constraints.

\section*{Acknowledgment}
The authors would like to thank the MICCAI 2018 IVDM3Seg and MICCAI 2017 MMWHS organizers for providing the data.

\bibliography{Refs}

\end{document}